%% file: online_bootstrap.tex
\documentclass{article}
\usepackage{preprint}
\usepackage{graphicx}
\usepackage{enumitem}

\usepackage{amssymb}
\usepackage{amsmath}
\usepackage{amsthm}
\usepackage{dsfont}
\usepackage{bm}
\usepackage{algorithm}
\usepackage{algpseudocode}
\usepackage{caption}
\usepackage{subcaption}
\usepackage{placeins}

\usepackage{hyperref}
\hypersetup{
	colorlinks   = true,
	urlcolor     = black,
	linkcolor    = black,
	citecolor   = black
}
\usepackage[capitalize,nameinlink]{cleveref}
\usepackage[round]{natbib}
\input{theorems-en.tex}

\input{math.tex}
\input{stats.tex}

\input{tex.tex}

\title{An Online Bootstrap for Time Series}
\date{}
\author{Nicolai Palm \\
LMU Munich \\ 
Munich Center for Machine Learning \\
\texttt{n.palm@lmu.de}\\
\AND
Thomas Nagler \\
LMU Munich \\ 
Munich Center for Machine Learning \\
\texttt{t.nagler@lmu.de}\\
}

\renewcommand{\headeright}{}
\renewcommand{\undertitle}{}

\begin{document}
\maketitle
\begin{abstract}
  Resampling methods such as the bootstrap have proven invaluable in the field of machine learning. 
  However, the applicability of traditional bootstrap methods is limited when dealing with large streams of dependent data, 
  such as time series or spatially correlated observations. In this paper, we propose a novel bootstrap 
  method that is designed to account for data dependencies and can be executed online, 
  making it particularly suitable for real-time applications.
  This method is based on an autoregressive sequence of increasingly dependent resampling weights.
  We prove the theoretical validity of the proposed bootstrap scheme under general conditions. 
  We demonstrate the effectiveness of our approach through extensive simulations 
  and show that it provides reliable uncertainty quantification even in the presence of complex data dependencies.
  Our work bridges the gap between classical resampling techniques and the demands of modern data analysis, 
  providing a valuable tool for researchers and practitioners in dynamic, data-rich environments.
\end{abstract}

\input{introduction.tex}
\input{background.tex}

\input{our-procedure.tex}

\input{experiments.tex}
\input{discussion.tex}

\bibliographystyle{apalike}
\bibliography{bibliography}
\input{supplement.tex}

\end{document}

%% file: theorems-en.tex
\usepackage{shadethm}
\usepackage{etoolbox}

\newtheoremstyle{new}
  {12pt}      % Space above
  {12pt}      % Space below
  {\itshape}  % Body font
  {}          % Indent amount (empty = no indent, \parindent = para indent)
  {\bfseries\color{black}} % head font
  {.}         % Punctuation after  head
  { }         % Space after  head: " " = normal inter-word space; \newline = line break
  {}          % head spec (can be left empty, meaning 'normal')
\theoremstyle{new}
\newtheorem{theorem}{Theorem}[section]
\newtheorem{corollary}[theorem]{Corollary}

\newtheorem{lemma}[theorem]{Lemma}
\newtheorem{definition}[theorem]{Definition}

\newtheorem{example}[theorem]{Example}

\newtheorem{construction}[theorem]{Construction}

\AfterEndEnvironment{theorem}{\noindent\ignorespaces}
\AfterEndEnvironment{corollary}{\noindent\ignorespaces}
\AfterEndEnvironment{proposition}{\noindent\ignorespaces}
\AfterEndEnvironment{lemma}{\noindent\ignorespaces}
\AfterEndEnvironment{definition}{\noindent\ignorespaces}
\AfterEndEnvironment{assumption}{\noindent\ignorespaces}
\AfterEndEnvironment{example}{\noindent\ignorespaces}
\AfterEndEnvironment{remark}{\noindent\ignorespaces}
\AfterEndEnvironment{claim}{\noindent\ignorespaces}

\Crefname{theorem}{Theorem}{Theorems}
\Crefname{corollary}{Corollary}{Corollaries}
\Crefname{proposition}{Proposition}{Propositions}
\Crefname{lemma}{Lemma}{Lemmas}
\Crefname{definition}{Definition}{Definitions}
\Crefname{example}{Example}{Examples}
\Crefname{remark}{Remark}{Remarks}
\Crefname{claim}{Claim}{Claims}

\usepackage[breakable, theorems, skins]{tcolorbox}
\tcbset{enhanced}

\definecolor{shadethmcolor}{cmyk}{0,0,0,0.075}    
\definecolor{shaderulecolor}{rgb}{1,1,1}   % color of frame 

\newtheoremstyle{shad}
  {12pt}      % Space above
  {12pt}      % Space below
  {\itshape }  % Body font
  {}          % Indent amount (empty = no indent, \parindent = para indent)
  {\bfseries\color{black}} % head font
  {.}         % Punctuation after head
  { }         % Space after head: " " = normal inter-word space; \newline = line break
  {}          % head spec (can be left empty, meaning 'normal')
\theoremstyle{shad} 

\newshadetheorem{thmbox}[theorem]{Theorem}
\newshadetheorem{corbox}[theorem]{Korollar}
\newshadetheorem{propbox}[theorem]{Proposition}
\newshadetheorem{lembox}[theorem]{Lemma}
\newshadetheorem{exbox}[theorem]{Beispiel}
\newshadetheorem{defbox}[theorem]{Definition}
\newshadetheorem{rembox}[theorem]{Anmerkung}

\newtheoremstyle{shad*}
  {12pt}      % Space above
  {12pt}      % Space below
  {\itshape }  % Body font
  {}          % Indent amount (empty = no indent, \parindent = para indent)
  {\bfseries\color{black}} % head font
  {.}         % Punctuation after head
  { }         % Space after head: " " = normal inter-word space; \newline = line break
  {}          % head spec (can be left empty, meaning 'normal')
\theoremstyle{shad*} 
\newshadetheorem{anobox}{}

\Crefname{theorem}{Theorem}{Theoreme}
\Crefname{corbox}{Korollar}{Korollare}
\Crefname{propbox}{Proposition}{Propositionen}
\Crefname{lembox}{Lemma}{Lemmas}
\Crefname{exbox}{Beispiel}{Beispiele}
\Crefname{defbox}{Definition}{Definitionen}
\Crefname{rembox}{Anmerkung}{Anmerkungen}

%% file: math.tex
\usepackage{bm}

\newcommand{\R}{\mathds{R}}
\newcommand{\N}{\mathds{N}}

\newcommand{\Z}{\mathds{Z}}
\newcommand{\E}{\mathds E}

\newcommand{\Ncal}{\mathcal{N}}
\newcommand{\Ocal}{\mathcal{O}}

\renewcommand{\bar}{\overline}

%% file: stats.tex
\providecommand{\Pr}{}
\renewcommand{\Pr}{\mathbb{P}}
\newcommand{\var}{{\mathds{V}\mathrm{ar}}}

\newcommand{\cov}{{\mathds{C}\mathrm{ov}}}

\newcommand{\wh}[1]{\widehat{#1}}

%% file: tex.tex
\makeatletter
\def\blfootnote{\gdef\@thefnmark{}\@footnotetext}
\makeatother

%% file: introduction.tex
\section{Introduction}

Uncertainty quantification (UQ) has become indispensable in statistics, machine learning, and numerous other
scientific disciplines. 
It plays a pivotal role in assessing the 
reliability of predictions, parameter estimates, and models. 
Bootstrapping is a universal ad-hoc approach for UQ and a cornerstone of many approaches leveraging UQ.
Especially in the context of theoretically unknown or difficult-to-compute uncertainty distributions, bootstrap methods have proven to be 
remarkably powerful.

A potential bottleneck in real applications is that computation of the
bootstrapped distributions and storage of underlying data gets expensive 
in time and memory with increasing amount of data. 
Especially in the context of big data sets and/or streaming data settings, 
this limits the applicability of standard bootstrap methods. 
Yet, this is a common setup in modern data analysis.

Online algorithms attempt to address these challenges by performing continuous, 
cheap updates of a model/estimate --- optimally processing 
only a fraction of the data within each iteration. This significantly 
decreases the associated costs 
and requires only a fraction of the data kept in memory. 

Existing bootstrap schemes, however, are either not computable by an online algorithm or make restrictive assumptions on dependence in the data.
This motivates the development of an online bootstrap scheme for general time series. 

Our main contributions can be summarized as follows:
\begin{enumerate}
    \item We propose a novel bootstrap procedure that (i) can be computed online and (ii) works for both independent and dependent data streams. To the best of our knowledge, this is the first such
    bootstrapping scheme.     
    \item We prove its theoretical validity under general conditions and provide theoretical insights into the optimal choice of hyperparameters. 
    \item We demonstrate validity and effectiveness through a number of simulations illustrating its advantages over the current state of the art.    
\end{enumerate}
The remainder of the paper is structured as follows. \cref{sec:background} provides some theoretical background and summarizes related work. \cref{sec:procedure} introduces the new method and presents the main theoretical results, which is evaluated empirically in \cref{sec:experiments}. \cref{sec:discussion} discusses applications and limitations. 
All proofs are provided in the supplementary material.

%% file: background.tex
\section{Background and related work} \label{sec:background}
\subsection{Online learning}

Online learning deals with problems where there is a continuous stream 
of data \citep{cesa2006prediction}. Such problems arise naturally 
if events are observed at the moment they occur. 
In other settings, a complete data set is available from the start, 
but it is computationally preferable to work through it sequentially or in batches. 
Online convex optimization methods like stochastic gradient descent are 
prime examples \citep{shalev2012online}. 

In such situations, one could recompute a quantity of interest at every time step 
using all data observed so far. However, this is very inefficient if 
the number of parameters or observations is large. 
In fact, re-computing on the full data set is often infeasible, 
because there is limited memory or limited time to update. For example, computing the sample average at $\bar X_n = \frac{1}{n} \sum_{t = 1}^n X_{t}$ naively requires $O(n)$ time at any given moment in time. 

Online algorithms strive to perform continuous 
updates that are cheap, optimally processing only a single observation 
at every step. 
Continuing the example above, the sample average at time $n$ adheres to the cheap update rule $\bar X_n = (1 - 1/n) \bar X_{n - 1} + X_{n} / n$, which scales as $O(1)$ in time and memory.
Such algorithms and their corresponding theoretical properties
are currently subject of interest in numerous fields,
such as optimization \citep{fang2018online,godichon2019online, zhong2023online}, 
multi-armed bandits \citep{dimakopoulou2021online,wan2023multiplier}
and reinforcement learning \citep{ramprasad2022online}.

\subsection{Bootstrapping}

Bootstrapping is considered as one of the fundamental achievements
of statistics \citep{kotz1992breakthroughs}.
Generally speaking, bootstrapping is a form of resampling:
From a given set of samples $x_1,\dots,x_n$ according to some random variables
$X_1,\dots,X_n$, we generate synthetic samples $x_1^*,\dots,x_n^*$, or more generally, synthetic 
random variables $X_1^*,\dots,X_n^*$
depending on $X_1,\dots,X_n$.
\citet{efron1992bootstrap} proposed the first bootstrap, 
based on sampling with replacement, nowadays called the empirical bootstrap.
\begin{example}[Empirical bootstrap]\label{ex:empiricial-bootstrap}
    The empirical bootstrap generates a new sample $X_1^*, \dots, X_n^*$ by drawing uniformly at random from the observed sample $\{X_1,...,X_n\}$ (with replacement).
\end{example}
The multiplier bootstrap \citep{vanderVaart1996} offers a general class 
of bootstrapping schemes based on perturbations of the 
original observations with suitable weights. 
The empirical bootstrap cannot be computed online, because it requires keeping track of the entire observed sample $\{X_1, \dots, X_n\}.$ The multiplier bootstrap does not suffer from this issue.

\begin{example}[Multiplier bootstrap]\label{ex:multiplier-bootstrap}
    Let $V_1, \dots, V_n$ be iid real valued random variables with 
    $$\E(V_i)=\var(V_i)=1.$$
    Then we obtain the multiplier bootstrap for iid data by $$X_i^*=\frac{V_i}{\bar V_n}X_i, \quad \text{where } \quad \bar V_n = \frac 1 n \sum_{i = 1}^n V_i.$$
    \vspace*{-24pt}
\end{example}
Popular special cases include the Gaussian bootstrap \citep{BURKE1998697} and Bayesian bootstrap \citep{rubin19891} where the weights $V_i$ are drawn from the standard normal and standard exponential distributions, respectively. 
The \emph{iid} multiplier bootstrap can be computed online.
Indeed, each $X_i^*$ depends only on the $i$-th observation $X_i$, a new weight $V_i$ and a running average of $V_i$'s. 
\citet{fang2018online} and \cite{zhong2023online} used this insight to design an online method for computing bootstrap confidence intervals for the SGD estimator. However, this method is only valid for \emph{iid} data.

Recall the random variables $X_i^*$ depend on $X_i$. We abbreviate the probabilities and variances given 
realizations of $X_i$ by
\begin{align*}
    \Pr^*(\cdot)&=\Pr(\cdot \mid X_1,X_2, \dots), \\
   \var^*(\cdot)&=\var( \cdot \mid X_1, X_2, \dots).
\end{align*}
These quantities depend on $X_1,...,X_n$ and are, therefore, itself random variables.
Now if the average over $(X_i)_{i\in \N}$ satisfies a central limit theorem, then, we expect the average over synthetic 
random variables to satisfy a similar central limit theorem. 

\begin{definition}[Bootstrap consistency]\label{def:consistency}
    Let $(X_i)_{i\in \N}$ and $(X_i^*)_{i\in \N}$ be sequences of $\R^d$-valued random variables and $X_i^*$ depending on $X_i$.
    Define 
    \begin{align*}
        T_n=\frac{1}{n}\sum_{i=1}^nX_i
        \text{ and }T_n^*=\frac{1}{n}\sum_{i=1}^nX_i^*
    \end{align*}
    The sequence $(X^*_i)_{i\in \N}$ is a \emph{consistent resampling scheme for $(X_i)_{i\in \N}$} if 
    \begin{align*}
        \sup_{x\in \R^d}&\bigl|\Pr^*\left\{\sqrt{n}(T_n^*-T_n) \leq x \right\} -\Pr\left\{\sqrt{n}(T_n-\E(T_n)) \leq x \right\}\bigr| \stackrel{n\to\infty}\to 0,
    \end{align*}
    in probability with respect to  $(X_i)_{i\in \N}$.
\end{definition}

In practice, one generates a few hundred resampled data sets and approximates the distribution $\Pr^*\left\{\sqrt{n}(T_n^*-T_n) \leq x \right\}$ by empirical quantities of the bootstrap replicates.
From this, other measures of uncertainty, such as confidence intervals 
or mappings of the underlying statistic, 
can be derived
\citep[e.g.,][Chapter 23]{van2000asymptotic}.

\Cref{ex:empiricial-bootstrap} and \Cref{ex:multiplier-bootstrap}
are consistent resampling schemes provided that 
the underlying data is independent, but fail otherwise.

\subsection{Time series bootstrap}

\citet{kunsch1989jackknife} proposed the blockwise bootstrap
as a general resampling scheme for time series.
Roughly, the idea is to draw overlapping blocks of observations. 
Since observations appear in blocks, the resampled observations naturally inherit 
dependencies from the original samples. 
Increasing the block lengths with the sample sizes make the blockwise bootstrap 
a consistent resampling scheme.

Later, extensions \citep{politis1993nonparametric} and 
further investigations \citep{hall1995} of the blockwise bootstrap arose.
Recently, \citet{liu2023statistical} use the blockwise bootstrap for inference in SGD estimators.
\cite{buhlmann1993blockwise} proposed a multiplier variant of the blockwise bootstrap that includes regular block bootstrap methods as a special case.
Here, the multiplier weights are itself a dependent time series. 
Increasing the multiplier weights' serial dependence then allows to capture the dependencies in the original observations.  

The above methods are not fit for the online setting, however.
For the method to work, all blocks have to increase in size with $n$.
To compute the bootstrap in practice,
the entire data set $(X_i)_{i = 1}^n$ needs to be kept in memory and processed fully, every time the block size changes.
This quickly becomes prohibitively expensive when $n$ is large.

%% file: our-procedure.tex
\section{New bootstrap procedure} \label{sec:procedure}

\subsection{Proposed method}\label{sec:our-methodology}

\Cref{ex:multiplier-bootstrap} gives rise to a general class of bootstrapping schemes by constructing synthetic random variables
$$X_i^*=\frac{V_i}{\bar V_n}X_i.$$
A key insight from the block multiplier bootstrap is the following: to remain valid for time series $(X_i)_{i \in \N}$, the
dependencies between weights $V_i$ and $V_j$ must increase with
the sample size $n$, but at the same time remain almost independent when the time gap $|i - j|$ is sufficiently large compared to $n$.
In the non-iid case, a scaling of the weights by their arithmetic mean is also necessary.

As a general construction of such $(V_i)_{i\in \N}$
we propose the following autoregressive sequence of weights:

\begin{construction}\label{ex:construction}
    Let $(\zeta_i)_{i\in \N}$ be an iid sequence such that $\zeta_i\sim \mathcal{N}(0,1)$.
    Define
    \begin{align*}
        V_0     & =0                                        \\
        V_{i} & =1+\rho_i(V_{i-1}-1)+\sqrt{1-\rho_i^2}\zeta_{i}
    \end{align*}
    where $\rho_i=1-i^{-\beta}$, $\beta \in (0,\frac{1}{2})$, and the bootstrapping scheme 
    $$X_i^*=\frac{V_i}{\bar{V}_n}X_i \quad \text{with} \quad \bar{V}_n=\frac 1 n\sum_{i=1}^nV_i.$$
    \vspace*{-24pt}
\end{construction}
The proposed weight sequence $(V_i)_{i \in \N}$ and the corresponding bootstrap average 
$$\bar X_n^* = \frac 1 n \sum_{i = 1}^n \frac{V_i}{\bar{V}_n} X_i$$ 
can naturally be computed with cheap online updates.
Indeed, the $i$-th bootstrapped variable $X_i^*=(V_i/\bar{V}_i)X_i$ depends only
on the predecessor weight $V_{i-1}$, a single random perturbation $\zeta_{i}$ and the scaling average $\bar{V}_{n}$.
The latter can be obtained recursively by 
$$\bar{V}_{n}=\frac{ (n - 1)\bar{V}_{n-1}+ V_n}{n}.$$
In summary, the bootstrap average can be updated via the relation 
$$\bar X_n^* = \frac{(n-1)\bar{V}_{n-1}\bar{X}_{n-1}^*+X_nV_n}{(n-1)\bar{V}_{n-1}+V_n},$$
which scales as $O(1)$ in memory and time.
This is a huge computational advantage compared to (multiplier) block bootstrap methods and opens new application areas for bootstrapping methods in online settings.

\begin{algorithm}
    \caption{Online AR-bootstrap}
    \label{alg:1}
    \textbf{Initialize:} $\bar X^{*(b)} = 0$, $V^{(b)} = 0$, $\bar{V}^{(b)} = 0$, $b = 1, \dots, B$. \\[11pt]
    \textbf{For times $t = 1, 2, \dots$}:
    \begin{enumerate}
        \item Observe new datum $X_{t}$.
        \item \textbf{For all $b = 1, \dots, B$}:
              \begin{enumerate}[label=(\roman*)]
                  \item Simulate $\zeta^{(b)} \sim \Ncal(0, 1)$.
                  \item With $\rho = 1 - t^{-\beta},$ update (in this order)
                  \begin{align*}
                    V^{(b)} &\leftarrow  1+\rho(V^{(b)}-1)+\sqrt{1-\rho^2}\zeta^{(b)}, \\
                    \bar X^{*(b)} &\leftarrow \frac{(t-1)\bar{V}^{(b)}\bar{X}^{*(b)}+X_tV^{(b)}}{(t-1)\bar{V}^{(b)}+V^{(b)}}, \\
                    \bar{V}^{(b)} &\leftarrow  (1-1/t)\bar{V}^{(b)}+V^{(b)} / t.
                \end{align*}
              \end{enumerate}
        \item Compute empirical variance and/or quantiles of $\{\bar X^{*(1)}, \dots, \bar X^{*(B)}\}$ to quantify uncertainty in $\bar X_t$.
    \end{enumerate}
\end{algorithm}

To quantify uncertainties in practice, we have to keep several independent bootstrap `chains' $\bar X_n^{*(1)}, \bar X_n^{*(2)},  \dots$.
From those, we can compute the empirical standard deviation or quantiles at any point in time.
The whole procedure is summarized in \cref{alg:1}.

\subsection{Theory}

In the following, we provide rigorous mathematical guarantees for the validity of the proposed scheme.
We first recall some common concepts from time series analysis.

\begin{definition}[Stationarity]
    A stochastic process $(X_i)_{i\in \N}$ is strictly stationary if $$P_{(X_{t_1},\dots,X_{t_n})}=P_{(X_{t_1+\tau},\dots,X_{t_n+\tau})}$$
    for every $\tau,n,t_1,\dots,t_n\in \N$.
\end{definition}
A stationary time series does not change its fundamental behavior, at least on large time scales. 
Stationarity is a standard condition for statistical limit theorems.
In applications, it is often ensured by appropriate pre-processing steps 
like detrending or differencing \citep[see,][and the discussion in \cref{sec:discussion}]{hamilton2020time}.

\begin{definition}[$\alpha$-mixing]
    Let $(X_i)_{i\in \N}$ be a strictly stationary stochastic process.
    Define the $\alpha$-mixing coefficient of order $h$
    \begin{align*}
        \alpha(h)=\sup \{
         & |\Pr(A\cap B)-\Pr(A)\Pr(B)|\colon s\in \N,
        A\in \sigma(X_i|i\leq s),B\in \sigma(X_i|i>s+h)\}.
    \end{align*}
    Then, $(X_i)_{i\in \N}$ is $\alpha$-mixing (or strong mixing) if
    $$\alpha(h)\stackrel{h\to \infty}\to 0.$$
    \vspace*{-24pt}
\end{definition}
The $\alpha$-mixing coefficient quantifies
how quickly the influence of past events diminishes as one
moves further into the sequence.
Accordingly, $\alpha$-mixing means that the events become close to independent when they are far apart in time.
If the strong mixing coefficients converge fast enough to zero,
the sequence satisfies a central limit theorem \citep[Theorem 1.7]{bosq2012nonparametric}.

Now denote by $$C^X(h)=\cov(X_i,X_{i+h})$$ the covariance of $X_i$ and $X_{i+h}$ for $h\geq 0$.
Since $(X_i)_{i\in \N}$ is stationary, $C^X(h)$ is independent of $i$. Our main results require the following conditions on the observed sequence $(X_i)_{i = 1}^n$.
\begin{enumerate}[label=(A\arabic*)]
    \item \label{A1} $\E(X_i^{8})<\infty$.
    \item \label{A2} $\alpha(i) = O(i^{-\gamma})$ for some $\gamma > 2$.
    \item \label{A3} $\lim_{n\to \infty}\frac{1}{n}\sum_{h=-n}^n |h|^{\frac{1}{\beta}} |C^X(|h|)|=0$.
\end{enumerate}
The first condition excludes extremely heavy tails in the variables $X_i$, the other two restrict the strength of dependence in the time series. 
Overall, the conditions should be considered rather mild in view of the time series bootstrap literature \citep[cf.,][]{kunsch1989jackknife,buhlmann1993blockwise}.

To simplify our asymptotic analysis of the procedure, we first have a closer look at the role of the scaling average $\bar V_n$. 
The following result shows that scaling by $\bar V_n$ implicitly allows to assume that $\E[X_i] = 0$, but is otherwise negligible asymptotically.
\begin{lemma} \label{lem:vn-negligible}
    If the time series $X_1, X_2, \ldots \in \R$ satisfies assumptions \ref{A1}--\ref{A3}, it holds 
    \begin{align*}
        &\quad \frac{1}{n}\sum_{i = 1}^n \frac{V_i}{\bar V_n} X_i - \frac{1}{n} \sum_{i = 1}^n X_i  
        = \frac{1}{n}\sum_{i = 1}^n V_i (X_i - \E[X_i]) - \frac{1}{n}\sum_{i = 1}^n  (X_i - \E[X_i]) 
        + o_P(n^{-1/2}).
    \end{align*}
    \vspace*{-24pt}
\end{lemma}

Our main result establishes the validity of the bootstrap scheme.
\begin{theorem}\label{thm:sequence-consistent}
    If the time series $X_1, X_2, \ldots \in \R$ satisfies assumptions \ref{A1}--\ref{A3}, 
    \Cref{ex:construction} provides a consistent resampling scheme. 
\end{theorem}
As a major part within the proof of the above theorem, we show that the proposed bootstrap procedure gives 
consistent estimates of the variance.
This investigation provides fundamental insights into the influence of the procedure's hyperparameter $\beta$.
In particular, we determine an optimal bias-variance trade-off.

\begin{theorem}\label{thm:variance-details}
    Define the target variance $$\sigma_\infty^2 = \lim_{n \to \infty}\var\left(\frac{1}{\sqrt{n}}\sum_{i = 1}^n {X_i}\right).$$
    If $\E[X_i] = 0$ and assumptions \ref{A1}--\ref{A3} hold, 
    \begin{enumerate}[label=(\alph*)]
        \item $\displaystyle  \E\biggl[\var^*\left(\frac{1}{\sqrt{n}}\sum_{i=1}^n V_iX_i\right)\biggr] -\sigma_\infty^2 = \mathcal{O}\bigl(n^{-\frac{\beta}{1+\beta}}\bigr),$
        \item $\displaystyle  \var\biggl[\var^*\left(\frac{1}{\sqrt{n}}\sum_{i=1}^n V_i X_i\right)\biggr]= \mathcal{O}\bigl(n^{\beta-1}\bigr),$
        \item $\displaystyle\var^*\left(\frac{1}{\sqrt{n}}\sum_{i=1}^n V_iX_i\right)\overset{P}{\to}\sigma_\infty^2$,
        \item the (asymptotically) optimal $\beta$ minimizing
        $$\E\biggl[\var^*\left(\frac{1}{\sqrt{n}}\sum_{i=1}^n V_i  X_i\right)-\sigma_\infty^2\biggr]^2$$
        is given by $$\beta_{opt}=\sqrt{2}-1.$$
    \end{enumerate}
    
\end{theorem}
In \cref{sec:proofs-avar}, we prove this result for a larger class of bootstrap weights $V_i/\bar{V}_n$ (see \Cref{cor:var-convergence}) under more involved assumptions. 
These assumptions are verified for our specific scheme from \cref{ex:construction} in \cref{lem:ass-cons-var-example}.
Note also that we omitted the scaling average $\bar V_n$ in this result.

The optimal choice $\beta_{opt}$ makes the mean-squared-error in the first display of (c) converge at rate $O(n^{\sqrt{2} - 2}) \approx O(n^{-0.59})$. 
This is slightly slower than the rate $O(n^{-2/3})$ attained by the blockwise bootstrap \citep[][Section 3.3]{buhlmann1993blockwise}. 
This is the statistical price we pay for the computational advantage of autoregressive bootstrap weights. 
In many applications, the latter can easily outweigh the small loss in statistical efficiency, see our experiments in \cref{sec:experiments}.

\subsection{Beyond the simple sample average}

The preceding results were stated for simple sample averages $n^{-1} \sum_{i = 1}^n X_i$ of real-valued random variables to simplify the exposition. 
The methodology is much more broadly applicable, however. 

\paragraph{Transformed random variables.} 
The most immediate generalization results from a simple relabeling. 
Suppose there is a sequence of random variables $(Z_i)_{i \in \N}$ and some function $f$. To quantify uncertainty in the statistic 
$$T_n = \frac 1 n \sum_{i = 1}^n f(Z_i),$$
we can use its bootstrapped version 
$$T_n^* = \frac 1 n   \sum_{i = 1}^n \frac{V_i}{\bar{V}_n} f(Z_i).$$
All results of the \cref{sec:our-methodology} apply naturally upon defining $X_i := f(Z_i)$.  
This also reveals the following, more fundamental interpretation of the multiplier bootstrap: 
each observation is assigned a weight $V_i/\bar{V}_n$, and these weights are used when computing bootstrapped averages --- irrespective of what exactly we're averaging. 

\paragraph{Multidimensional vectors.}
\cref{thm:sequence-consistent} immediately extends to averages of random vectors through the Cram\'er-Wold device \citep[e.g.,][p.~16]{van2000asymptotic}.

\begin{corollary} \label{cor:multidimensional}
Let $Z_1, Z_2, \ldots \in \R^d$
 and assume that \ref{A1}--\ref{A3} hold for $X_i := \gamma^\top Z_i$ and every $\gamma \in \R^d$. Then \cref{ex:construction} provides a consistent resampling scheme.
 \vspace*{-12pt}
\end{corollary}

A similar generalization of \cref{thm:variance-details} also holds for vectors, but is omitted for brevity.

\paragraph{Transformations of the sample average.}
The delta method for bootstrap (e.g. Theorem 23.5 in \cite{van2000asymptotic})
generalizes the consistency results of \cref{thm:sequence-consistent} 
to transformations of the sample average.
\begin{corollary}\label{cor:delta}
    Let $\phi:\R^d\to \R^k$ be continuously differentiable and assume that 
    \ref{A1}--\ref{A3} hold for $\gamma^TX_i$ with $X_i\in\R^d$ and all $\gamma\in \R^d$. 
    Then,     
    \begin{align*}
        \sup_{x\in \R^d}&\bigl|\Pr^*\left\{\sqrt{n}(\phi(T_n^*)-\phi(T_n)) \leq x \right\} 
        -\Pr\left\{\sqrt{n}(\phi(T_n)-\phi(\E(T_n))) \leq x \right\}\bigr| \stackrel{n\to\infty}\to 0,
    \end{align*}
    in probability with respect to  $(X_i)_{i\in \N}$.
\end{corollary}
This enables a broader application of the proposed 
methodology, e.g. to the sample variance (Example 23.6. in \cite{van2000asymptotic}).

%% file: experiments.tex
\section{Numerical validation} \label{sec:experiments}

In this section, we assess the performance and 
applicability of the proposed bootstrap via simulations.
In particular, we verify the theoretical results in a finite sample setting, illustrate the necessity of tailored bootstrap schemes for time series, and the computational benefits of our new method.
\begin{figure*}[h]
    \centering
    \resizebox{0.9\textwidth}{!}{\input{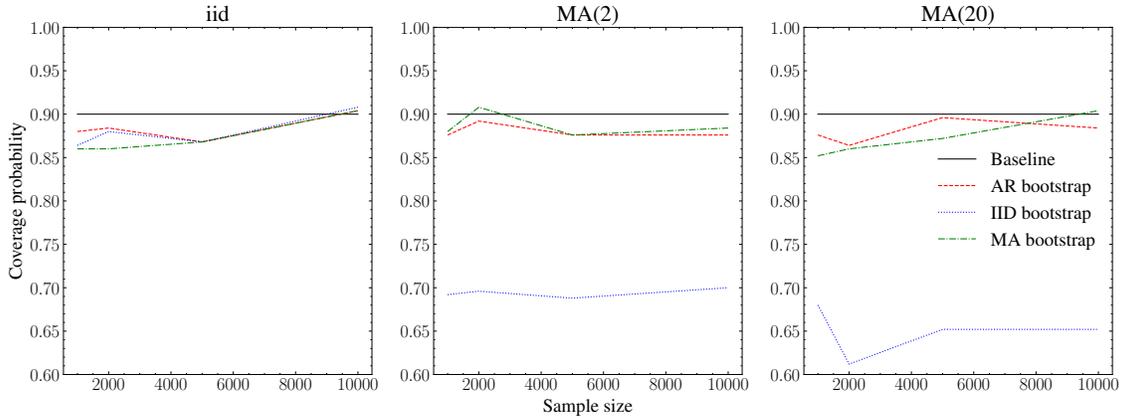}}
    \caption{Estimated coverage probability of the bootstrap procedures. The target level of $90\%$ is shown as solid line.}
    \label{fig:coverage}
\end{figure*}
\begin{figure*}[h]
    \centering
    \resizebox{0.9\textwidth}{!}{\input{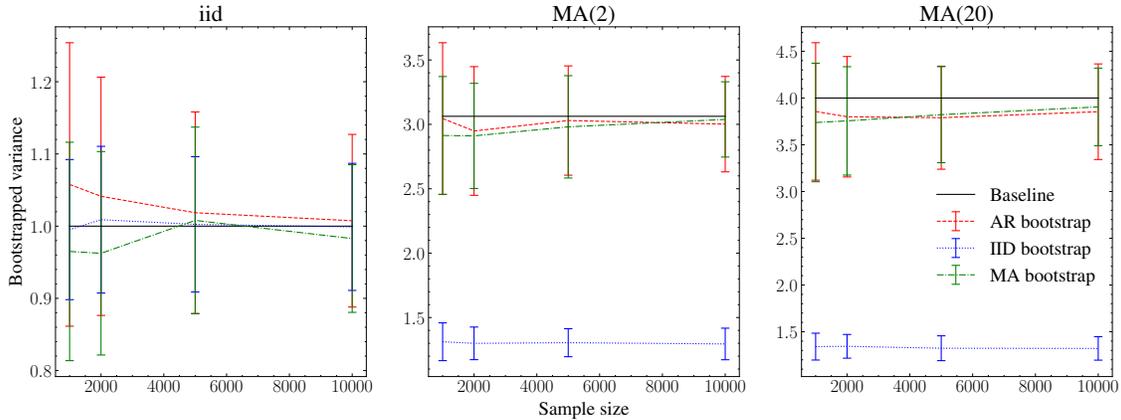}}
    \caption{Average plus/minus standard deviation of the estimated variances. The target level $\sigma_\infty$ is shown as solid line.}
    \label{fig:accuracy}
\end{figure*}

\subsection{Experimental design}
\paragraph{Data generating processes.}
We simulate $X_i$ from a moving average process of order $q\in \N$, $MA(q)$ for short, i.e., a stochastic process of the form
\begin{align*}
    X_i&=\mu + \epsilon_i+\sum_{j=1}^q\theta_j\epsilon_{i-j},
\end{align*}
with model parameters $\theta_1,...,\theta_q\in \R$ and \emph{iid} noise $\epsilon_i\sim \mathcal{N}(0,1)$.
%%%
In addition, we simulate from a nonlinear transformation of such a process 
$$Y_i=\exp(X_i)$$
and consider a nonlinear function of the sample averages
$$\ln\bigl[n^{-1}\sum_{i=1}^{n}\exp(X_i)\bigr]$$
for $q=2$ (referred to as $LogMeanExp$). 
The bootstrapped distribution reflects the correct distribution according to \Cref{cor:delta}.
Further, we simulate from 
a nonlinear process $MA(2)-GARCH(1,1)$ reflecting the volatility clustering typical 
for financial time series. 
We realize the latter by
\begin{align*}
    Z_i &= \mu+Z_i+\theta_1\gamma_{i-1}+\theta_2\gamma_{i-2}
\end{align*}
for $\gamma_i=\sigma_i\xi_i$,
$\sigma_i^2=\alpha_0+\alpha \gamma_{i-1}^2+\beta \sigma_{i-1}^2$ and $\xi_{i}\sim \mathcal{N}(0,1)$ iid.
%%%
\begin{figure*}[t!]
    \centering
    \includegraphics[width=0.6\textwidth]{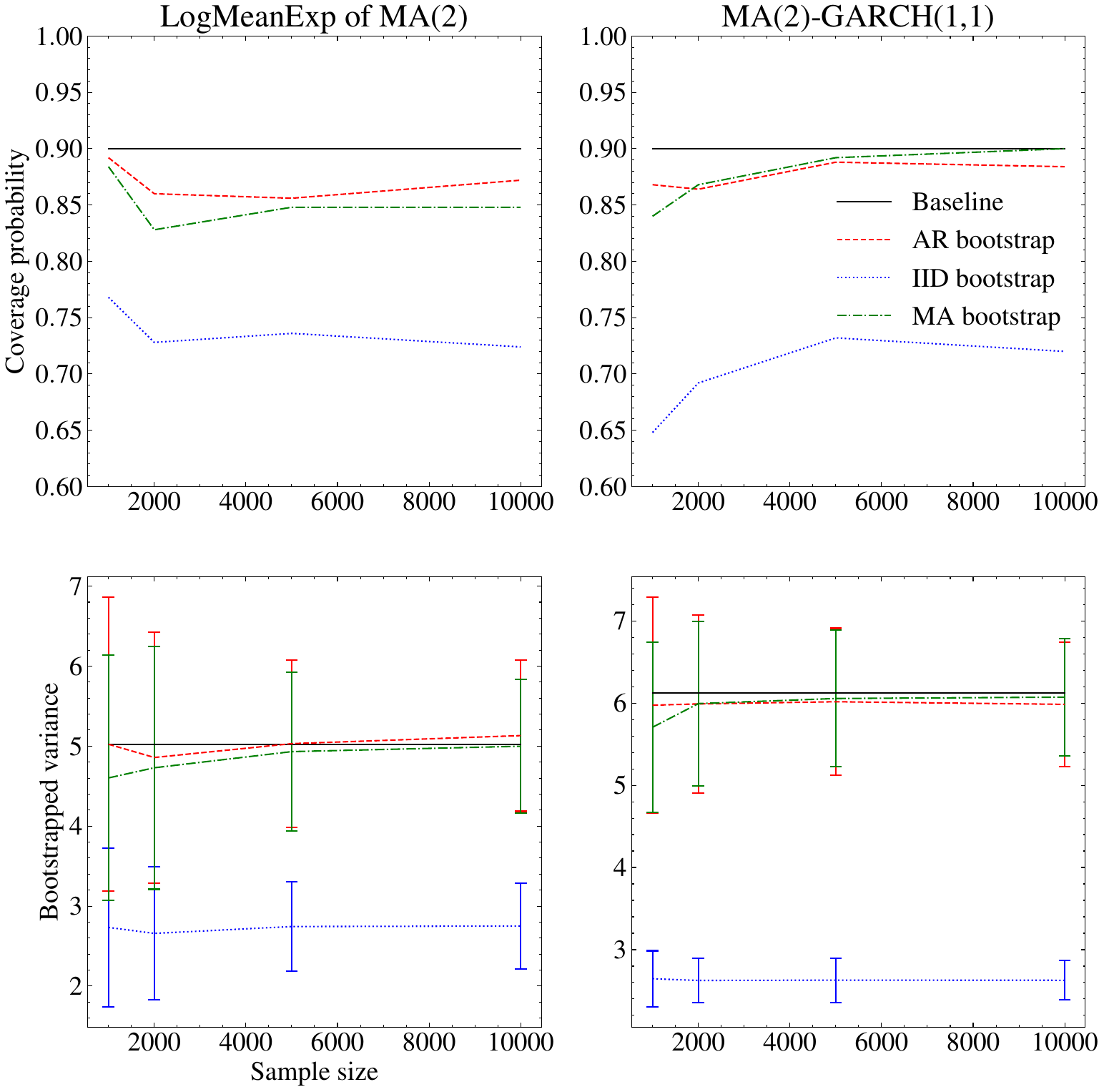}    
    \caption{Estimated coverage probability of the bootstrap procedures with target level of $90\%$ shown as solid
    line (top) and average plus/minus standard deviation of the estimated variances with target level $\sigma_\infty$ shown as
    solid line (bottom).}
    \label{fig:non-linear}
\end{figure*}

All processes, except the latter, are $q$-dependent\footnote{$(X_i)_{i \in \N}$ is called $q$-dependent if
 $(\dots, X_{i - 1}, X_i)$ is independent from $(X_{i + q+1}, X_{i + q + 2}, \dots)$ for every $i$. Then, $$C^X(|h|),\alpha(i)=0$$ for all $i,|h|>q$, from which assumptions (A2) and (A3) follow immediately.} 
and, hence, satisfy the assumptions of our theoretical results. 
For the latter we refer to \cite{lindner2009stationarity}.
We set $\theta_j = 2^{-j}$ and consider five scenarios:
\begin{itemize}
    \item $MA(0)$ corresponds to the \emph{iid} setting,
    \item $MA(2)$ corresponds to short-term serial dependence,
    \item $MA(20)$ corresponds to medium-term serial dependence.
    \item $LogMeanExp$ corresponds to a transformation of a nonlinear process with short-term serial dependence.
    \item $MA(2)-GARCH(1,1)$ corresponds to a nonlinear stochastic volatility process.
\end{itemize}

\paragraph{Bootstrap methods.}

We apply three bootstrap procedures of the form 
$$X_i^*=\frac{V_i}{\bar{V}_n}X_i$$ with
\begin{enumerate}[label = (\roman*)]
    \item  $V_i\sim\mathcal{N}(1,1)$ corresponding to \Cref{ex:multiplier-bootstrap} (\texttt{IID bootstrap});
    \item $V_i$ according to the moving average block bootstrap of \citet{buhlmann1993blockwise}, see \Cref*{ex:ma-bootstrap-buehlman} in the supplementary material, with $m_n=\lfloor n^{1/3}\rfloor$ (\texttt{MA bootstrap});
    \item $V_i$ according to our new \Cref{ex:construction} with parameter $\beta=\sqrt{2}-1$ (\texttt{AR bootstrap}).
\end{enumerate} 

\paragraph{Evaluation.}

For each simulated time series and method, we generate 250 bootstrap samples and compute the sample variance and a 90\%-confidence interval.
We repeat this procedure $M=250$ times.  
We assess the performance by a) mean and standard deviation of the estimated asymptotic variance $\sigma_\infty$, b) coverage probability 
of the resulting confidence interval, c) computation time.
See \Cref*{ap:evaluation-metrics} for computational details of the 
evaluation metrics.
The corresponding source code is provided at \url{https://github.com/nicolaipalm/online-bootstrap-implementation}.

\subsection{Results}

\paragraph{Validity.}

We start by checking the validity of confidence intervals constructed from the various bootstrap methods.
\Cref{fig:coverage} and the top panels of \Cref{fig:non-linear} 
plot the respective coverage probabilities against the sample size.
The target level of $90\%$ is indicated by the solid line.
When the data is generated as an \emph{iid} sequence (left panel of \Cref{fig:coverage}), all three bootstrap variants have approximately correct coverage, especially for large samples.
When there is dependence in the data (remaining panels), the \texttt{IID bootstrap} fails catastrophically, however. 
This is even the case for the linear MA(2) scenario, where dependence is weak and short-term (middle panel of \Cref{fig:coverage}). 
The two dedicated time series bootstraps (\texttt{MA} and \texttt{AR}) achieve approximately correct coverage in all five scenarios, 
even in the presence of nonlinear dependencies and transformations of the sample average (top panels of \Cref{fig:non-linear}).

\paragraph{Accuracy.}

We now dive deeper into how well the bootstrap methods estimate the true variance $\sigma_\infty$. 
\Cref{fig:accuracy} and the bottom panels of \Cref{fig:non-linear} show average estimates plus/minus their standard deviation and the target level.
Unsurprisingly, the \texttt{IID} bootstrap works best when the data is actually an independent sequence (left panel of \Cref{fig:accuracy}). 
In particular, it has virtually no bias, and the smallest variance among all methods.
In the time series settings it fails. 
On the other hand, the two time series bootstraps approach the target level, with bias and variance decreasing with the sample size
in all five scenarios.
The \text{MA} bootstrap appears to have a slightly smaller variance compared to our new \text{AR} method.
This reflects the small statistical cost we pay for its computational advantage, see our comments after \cref{thm:variance-details}.

\paragraph{Computation time.}
 
\begin{figure}[t!]
    \centering
      \centering
      \resizebox{0.45\textwidth}{!}{\input{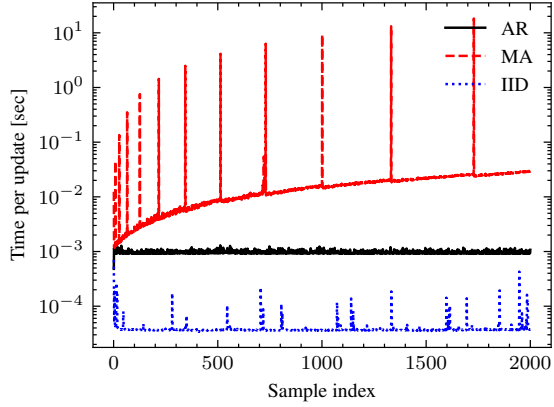}}
      \captionof{figure}{Computation time per online update of 200 bootstrap samples as the algorithms progress through a stream of samples.}
      \label{fig:time}
\end{figure}

The true benefit of the newly proposed scheme is the ability to compute it with cheap online updates. \Cref{fig:time} shows the computation time of an update step when the three bootstrap methods are used to generate 250 bootstrap samples in an online setting.
We see that \texttt{AR} and \texttt{IID} require a small, constant amount of time for every update as the algorithms progress.
The \texttt{IID} bootstrap is fastest, but invalid for time series data.
The blockwise bootstrap \texttt{MA} allows for cheap online updates as long as the block size remains constant. 
It occurs a huge cost whenever the block size needs to be increased: one must regenerate all past and current bootstrap weights and recalculate the bootstrap averages with new weights.
This shows as large spikes in \Cref{fig:time}.
Additionally, the time it requires quickly increases with time.
The last block update at around just 1700 samples already takes 20 seconds, where the other two methods remain in the milliseconds.
It is not reasonable to compute \texttt{MA} on much longer data streams.
Our new \texttt{AR} bootstrap on the other hand remains fast and valid.

%% file: discussion.tex
\section{Discussion} \label{sec:discussion}

We close with a discussion of potential applications and current limitations of our method.

\subsection{Applications in machine learning}

\paragraph{Empirical risk minimizers.}
Consider a parametrized prediction model $f_{\theta}$, a loss function $L$ and the empirical risk minimizer
\begin{align} \label{eq:erm}
  \wh \theta = \arg \min_{\theta} \sum_{i = 1}^n L(f_\theta, Z_i).
\end{align}
Under some regularity conditions, one usually has  \citep[see, e.g.,][]{giordano19a}
\begin{align} \label{eq:erm-approx}
  \wh \theta  - \theta_0 \approx -H^{-1} \frac{1}{n} \sum_{i = 1}^n \nabla_\theta L(f_{\theta_0}, Z_i),
\end{align}
where $\theta_0 = \arg \min_{\theta} \E[L(f_\theta, Z_i)]$ minimizes the true risk and $H = \E[\nabla_{\theta\theta} L(f_{\theta_0}, Z_i)]$ is the expected loss Hessian.
In machine learning, the uncertainty of predictions $f_{\wh \theta}(x)$ is more interesting than the parameter $\theta$.
A first-order Taylor approximation and \eqref{eq:erm-approx} give
\begin{align*}
  f_{\wh \theta} - f_{\theta_0} \approx  -\nabla_{\theta} f_{\theta_0} H^{-1} \frac{1}{n} \sum_{i = 1}^n \nabla_\theta L(f_{\theta_0}, Z_i).
\end{align*}

Now define the bootstrapped parameter
\begin{align*}
  \wh \theta^* = \arg \min_{\theta} \sum_{i = 1}^n \frac{V_i}{\bar{V}_n} L(f_\theta, Z_i),
\end{align*}
for which similar arguments yield
\begin{align*}
  f_{\wh \theta^*} - f_{\wh \theta} \approx -\nabla_{\theta} f_{\theta_0}(x) H^{-1} \frac{1}{n} \sum_{i = 1}^n \frac{V_i}{\bar{V}_n} \nabla_\theta L(f_{\theta_0}, Z_i).
\end{align*}
Applying \cref{cor:multidimensional} to the average on the far right of the last display now shows that the distribution of $f_{\wh \theta^*} - f_{\wh \theta}$ appropriately reflects the uncertainty of $f_{\wh \theta} - f_{\theta_0}$.

\paragraph*{SGD and online convex optimization.}

In practice, optimization problems like \eqref{eq:erm} are often solved using algorithms from online convex optimization, e.g., stochastic gradient descent.
In the \emph{iid} setting, it is well established that approximation \eqref{eq:erm-approx} also holds for averaged SGD updates \citep{ruppert1988efficient,polyak1992acceleration,fang2018online,zhong2023online}. Similar results for SGD on time series data were established recently by, e.g., \citet{godichon2019online,godichonbaggioni2023learning,liu2023statistical}.

\paragraph{Bandit algorithms.}

Our new bootstrap scheme can also be incorporated into bandit algorithms, similar to \citet{wan2023multiplier} in the \emph{iid} case.
In a simple multi-armed bandit, an agent picks some arm $A_t \in \{1, \dots, K\}$ and receives reward $R_{t, A_t}$ in return, at every time $t$. 
If $S_a = \{t\colon A_t = a\}$ is the set of times action $a$ was played, the expected reward of arm $k$ can be estimated as a simple sample average
\begin{align*}
  \wh r_a = \frac{1}{|S_a|}\sum_{t \in S_a} R_{t, a}.
\end{align*}
\citet{wan2023multiplier} proposed to use an independent multiplier bootstrap to quantify the uncertainty about $\wh r_a$, and use this to guide the exploration/exploitation-trade-off of the algorithm.
With our new method, the assumption that rewards $R_{t, a}$ are independent can be relaxed.

\subsection{Limitations}

\paragraph*{Stationarity assumption.}

As mentioned in \cref{sec:our-methodology}, stationarity of the series $(X_i)_{i \in \N}$ is a common assumption in the time series literature.
In applications, stationarity is often ensured by pre-processing steps. 
Technically, these steps should also be accounted for in uncertainty quantification, but this is difficult to do with generality.
Another way to alleviate this issue would be to extend our results to averages of nonstationary series.
Some recent developments in this field \citep{MERLEVEDE2020108581} can likely be adapted in future work.

\paragraph*{Negative weights.} 
In \cref{ex:construction}, we explicitly defined the weights $V_i$ to follow a Gaussian AR-process.
This choice is motivated by mathematical convenience: by construction, the distribution of the bootstrapped average (conditional on the data) is normal --- no central limit theorem is required.
A downside is that the bootstrap weights $V_i$ can be negative.
When the $X_i$'s are positive variables (counts, lengths, prices, etc.) and the sample size is small, this may be problematic.
The bootstrapped average $n^{-1}\sum_{i = 1}^n \frac{V_i}{\bar{V}_n} X_i$ could become negative, which results in meaningless estimates.
While this is not an issue asymptotically, it might be preferable to work with a strictly positive sequence $(V_i)_{i = 1}^n$. 
To prove the validity of such schemes would then require new central limit theorems for time series with increasing dependence, which is an interesting problem for future work.

%% file: supplement.tex
\renewcommand{\thesection}{\Alph{section}}
\renewcommand{\thesubsection}{\alph{subsection}}

\newpage
\title{Supplementary Materials}
\appendix

\section{Proving consistency of bootstrapping schemes}
Proving a bootstrapping scheme to be consistent mostly proceeds in the following two steps
\begin{enumerate}
    \item[S1] Prove that the random variables $X_i$ satisfy some central limit theorem and
    \item[S2] prove that the bootstrapped random variables $X_i^*$ satisfy some central limit theorem with the same limit distribution.
\end{enumerate}
In practice, a lot of central limit theorems are already well established each covering different assumptions on the random variables $X_i$,
i.e. S1 is given.
For the sake of clarity, we assume $\E(X_i)=0$. 
\Cref{lem:vn-negligible} treats the non-centered case.

Observe that $$\var^V\left(\frac{1}{\sqrt{n}}\sum_{i=1}^nV_iX_i)\right)$$ is a random variable induced by $X_i$ where we write $\var^V=\var^*$ with respect to the explicit construct $X^*=VX$.
S2 implies 
\begin{align*}
    \var^V\left(\frac{1}{\sqrt{n}}\sum_{i=1}^nV_iX_i\right)\overset{n\to \infty}{\to} \sigma_\infty^2
\end{align*}
in probability on the sequence $(X_i)_{i\in \N}$ where $\sigma_\infty^2$ denotes the asymptotic (finite) variance 
$$\lim_{n\to \infty}\var\left(\frac{1}{\sqrt{n}}\sum_{i=1}^nX_i\right)=\sigma_\infty^2.$$
Classically, such results are derived by proving 
\begin{align*}
    \E^X\left[\var^V\left(\frac{1}{\sqrt{n}}\sum_{i=1}^nV_iX_i\right)\right]
    \overset{n\to \infty}{\to} \sigma_\infty^2 \qquad \text{and} \qquad 
    \var^X\left[\var^V\left(\frac{1}{\sqrt{n}}\sum_{i=1}^nV_iX_i\right)\right]\overset{n\to \infty}{\to} 0,
\end{align*}
and then using Chebyshev's inequality.
Without loss of generality assuming $\E(X_i)=0$ (see \cref*{lem:vn-negligible}), a straightforward calculation exhibits 
\begin{align*}
    \var\left(\frac{1}{\sqrt{n}}\sum_{i=1}^nX_i\right)&=\frac{1}{n}\sum_{i,j=1}^n\cov(X_i,X_j), \\
    \E^X\left[\var^V\left(\frac{1}{\sqrt{n}}\sum_{i=1}^nV_iX_i\right)\right]&=\frac{1}{n}\sum_{i,j=1}^n\cov(V_i,V_j)\cov(X_i,X_j).
\end{align*}
This suggests that we need $\cov(V_i,V_j)\approx 1$ for $i,j$ great enough and $\cov(X_i,X_j)\not\approx0$.
However, calculating
\begin{align*}
    \var^X\left[\var^V\left(\frac{1}{\sqrt{n}}\sum_{i=1}^nV_iX_i\right)\right]
        & = \frac{1}{n^2}\sum_{i_1,\dots,i_4=1}^n\cov(V_{i_1},V_{i_2})\cov(V_{i_3},V_{i_4})\cov(X_{i_1}X_{i_2},X_{i_3}X_{i_4})
\end{align*}
requires $\cov^V(V_i,V_j)\approx 0$ for sufficiently many $i,j$ to vanish.
In the next section we provide corresponding formal results about the asymptotic variance.
These steps will be worked out in detail in the following sections.

\section{Proof of \cref*{lem:vn-negligible}}

It holds 
\begin{align*}
     \frac{1}{n}\sum_{i = 1}^n \frac{V_i}{\bar V_n} X_i - \frac{1}{n} \sum_{i = 1}^n X_i 
     =  \frac{1}{n}\sum_{i = 1}^n \left(\frac{V_i}{\bar V_n} - 1\right) X_i 
    & =  \frac{1}{n}\sum_{i = 1}^n \left(\frac{V_i}{\bar V_n} - 1\right) (X_i  - \E[X_i])  + \E[X_1]\underbrace{\frac{1}{n}\sum_{i = 1}^n \left(\frac{V_i}{\bar V_n} - 1\right)}_{ = 0}.
\end{align*}
Now 
\begin{align*}
     \frac{1}{n}\sum_{i = 1}^n \left(\frac{V_i}{\bar V_n} - 1\right) (X_i  - \E[X_i])        &= \frac{1}{n}\sum_{i = 1}^n (V_i- 1) (X_i  - \E[X_i])
     + \frac{1}{n}\sum_{i = 1}^n \left(\frac{V_i}{\bar V_n} - V_i\right) (X_i  - \E[X_i]) \\
     &= \frac{1}{n}\sum_{i = 1}^n (V_i- 1) (X_i  - \E[X_i])
     + \left(\frac{1}{\bar V_n} - 1\right)\frac{1}{n}\sum_{i = 1}^n V_i (X_i  - \E[X_i]).
\end{align*}
We observe 
\begin{align*}
    \var(\bar{V}_n)
    = \frac{1}{n^2}\sum_{i,j=1}^n\cov(V_i,V_j)
    \le \frac{1}{n^2}\sum_{i=1}^n \sum_{h\in \Z} v(i,i+|h|)
    \overset{n\to \infty}{\to} 0,
\end{align*}
by (iv) of \Cref{lem:variance}. Chebychev's inequality then yields $\bar{V}_n\overset{P}{\to}\E(V_i)=1$ and, consequently,  $ 1/\bar V_n - 1 \stackrel{P}\to 0$.
Now note that 
\begin{align*}
    \E\left[\frac{1}{n}\sum_{i = 1}^n V_i (X_i  - \E[X_i])\right] =  \E^V\left[\frac{1}{n}\sum_{i = 1}^n V_i \E^X[(X_i  - \E[X_i])] \right] = 0.
\end{align*}
The law of total variance gives
\begin{align*}
    \var\left[\frac{1}{n}\sum_{i = 1}^n V_i (X_i  - \E[X_i]) \right] &=  \E^V \left[\var^X \left[\frac{1}{n}\sum_{i = 1}^n V_i (X_i  - \E[X_i]) \right] \right] +  \var^V \biggl[\underbrace{\E^X \left[\frac{1}{n}\sum_{i = 1}^n V_i (X_i  - \E[X_i]) \right]}_{= 0} \biggr] \\
    &= \E^V \left[\var^X \left[\frac{1}{n}\sum_{i = 1}^n V_i (X_i  - \E[X_i]) \right] \right] \\
    &=  \frac 1 {n^2} \sum_{i = 1}^n \sum_{j = 1}^n \E[V_i V_j] \cov(X_i, X_j) \\
    &\le  \frac 2 {n^2} \sum_{i = 1}^n \sum_{j = 1}^n |\cov(X_i, X_j)| \tag*{by Cauchy-Schwarz and $\E[V_i^2] = 2$}\\
    &\le   \frac 1 {n}  \sum_{h \in \Z} |C^X(h)| \tag*{using that $(X_i)_{i \in \N}$ is stationary} \\
    &= \mathcal{O}\left(\frac 1 n\right). \tag*{by assumption (A3)}
\end{align*}
Applying Chebyshev's inequality again, we have shown
\begin{align*}
    \left(\frac{1}{\bar V_n} - 1\right)\frac{1}{n}\sum_{i = 1}^n V_i (X_i  - \E[X_i]) = o_P(1) \times \Ocal_P(n^{-1/2}),
\end{align*}
and the claim follows. \qedsymbol

\section{Asymptotic variance} \label{sec:proofs-avar}

In the following, we consider $(X_i)_{i\in \N}$ to be a real-valued 
strictly stationary stochastic process and 
abbreviate $v(i,j)=\cov(V_i,V_j)$.

We frequently sum over two indices, i.e. we consider a sum of the form $\sum_{i,j=1}^na_{ij}$. 
The reader convinces himself that 
\begin{align*}
    \{(i,j)|i,j=1,\dots,n\}
\end{align*}
is the disjoint union of 
\begin{align*}
    \{(i,i+j)|j=0,\dots,n,i=1,\dots,n-j\}
    \text{ and } \{(i+j,i)|j=1,\dots,n,i=1,\dots,n-j\}.
\end{align*}
If we assume $a_{ij}=a_{ji}$, then, we identify 
\begin{align}\label{eq:sum}
    \sum_{i,j=1}^na_{ij}=\sum_{j=-n}^n\sum_{i=1}^{n-|j|}a_{i,i+|j|}.
\end{align}

\begin{lemma}\label{lem:expected-value}
Assume the following:
\begin{enumerate}[label=(\roman*)]
    \item $\E(X_i)=0$ for all $i$.
    \item $|v(i,j)|\leq C$ for some $C\in \R$ and all $i,j$.
    \item $\sum_{h=-\infty}^\infty |C^X(|h|)|<\infty$. 
    \item for all $\epsilon>0$ and $h$ such that $C^X(|h|)\neq 0$ there exists $n_{\epsilon,|h|}$ such that $$|1-v(i,i+|h|)| \le \epsilon$$ for all $i \ge n_{\epsilon,|h|}$ and 
    \item $\lim_{n\to \infty}\frac{1}{n}\sum_{h=-n}^n n_{\epsilon,|h|} |C^X(|h|)|=0$
\end{enumerate}
Then, the asymptotic variance 
$$\var\biggl(\frac{1}{\sqrt{n}}\sum_{i=1}^nX_i\biggr)\overset{n\to \infty}{\to} \sigma_\infty^2=\frac{1}{n}\sum_{i,j=1}^\infty\cov(X_i,X_j)$$ exists, 
$$\E^X\left[\var^V\left(\frac{1}{\sqrt{n}}\sum_{i=1}^n V_iX_i\right)-\sigma_\infty^2\right] = \mathcal{O}\biggl(\frac{1}{n}\sum_{h=-n}^nn_{\epsilon,|h|}|C^X(|h|)| + \epsilon\sum_{h=-n}^n|C^X(|h|)|\biggr)$$
and, in particular,
$$\E^X\left[\var^V\left(\frac{1}{\sqrt{n}}\sum_{i=1}^nV_iX_i\right)\right]\overset{n\to \infty}{\to} \sigma_\infty^2.$$
\end{lemma}

\begin{proof}
    We observe 
    $$\var\biggl(\frac{1}{\sqrt{n}}\sum_{i=1}^nX_i\biggr)=\sum_{i,j=1}^n\cov(X_i,X_j).$$
   Therefore, its asymptotic variance (if it exists) is given by 
    $$\var\biggl(\frac{1}{\sqrt{n}}\sum_{i=1}^nX_i\biggr)\overset{n\to \infty}{\to} \sigma_\infty^2=\frac{1}{n}\sum_{i,j=1}^\infty\cov(X_i,X_j).$$
    Furthermore, we calculate
    \begin{align*}
        \var\left(\frac{1}{\sqrt{n}}\sum_{i=1}^nX_i\right)&=\frac{1}{n}\sum_{i,j=1}^n\cov(X_i,X_j)
        \\
        &=\frac{1}{n}\sum_{h=-n}^n\sum_{i=1}^{n-|h|}\cov(X_i,X_{i+|h|})
        & \text{by \eqref{eq:sum}}
        \\
        &= \frac{1}{n}\sum_{h=-n}^n\sum_{i=1}^{n-|h|}C^X(|h|)
        & \text{by stationarity of $(X_i)_{i\in \N}$}
        \\
        &= \sum_{h=-n}^n\frac{n - |h|}{n}C^X(|h|).
        \\
        &\le  \sum_{h=-\infty}^\infty|C^X(|h|)|.
    \end{align*}
    Therefore, the asymptotic variance exists by (iii).

We calculate 
\begin{align*}
    \E^X\left[\var^V\left(\frac{1}{\sqrt{n}}\sum_{i=1}^n V_iX_i\right)\right] 
    &=\E^X\left(\frac{1}{n}\sum_{i,j=1}^nv(i,j)X_iX_j\right)
    \\
    &=\frac{1}{n}\sum_{i,j=1}^nv(i,j)\E(X_iX_j)
    \\
    &=\frac{1}{n}\sum_{i,j=1}^nv(i,j)\cov(X_i,X_j).
    & \text{by assumption (i)}
\end{align*}
Hence,
\begin{align*}
    \lim_{n\to \infty} \frac{1}{n}\sum_{i,j=1}^n(v(i,j)-1)\cov(X_i,X_j)
    &=
    \lim_{n\to \infty}  \E^X\left[\var^V\left(\frac{1}{\sqrt{n}}\sum_{i=1}^n V_iX_i\right)\right] -\sigma_\infty^2.
\end{align*}
By \eqref{eq:sum} and stationarity of $(X_i)_{i\in \N}$ we obtain 
\begin{align*}
    \frac{1}{n}\sum_{h=-n}^n\sum_{i=1}^{n-|h|}(v(i,i+|h|)-1)C^X(|h|)
    &= \frac{1}{n}\sum_{i,j=1}^n(v(i,j)-1)\cov(X_i,X_j).
\end{align*}
Therefore, we consider 
$$\frac{1}{n}\sum_{h=-n}^n\sum_{i=1}^{n-|h|}(v(i,i+|h|)-1)C^X(|h|)$$
and prove that it converges to zero for $n\to \infty.$

We split the sum 
\begin{align*}
    \sum_{i=1}^{n-|h|}(v(i,i+|h|)-1)C^X(|h|)=&\sum_{i=1}^{n_{\epsilon, |h|}}(v(i,i+|h|)-1)C^X(|h|)
    +
    \sum_{i=n_{\epsilon,|h|}+1}^{n-|h|}(v(i,i+|h|)-1)C^X(|h|)
\end{align*}
and prove that the absolute values of both summands converge to zero.
For fixed $h$ we calculate
\begin{align*}
\biggl| \sum_{i=n_{\epsilon,|h|}+1}^{n-|h|}(v(i,i+|h|)-1)C^X(|h|) \biggr|
& \le \sum_{i=n_{\epsilon,|h|}+1}^{n-|h|}\epsilon |C^X(|h|)|
& \text{by assumption (iv).}
\\
&=\epsilon(n-|h|-n_{\epsilon,|h|}-1)|C^X(|h|)|
\end{align*}
for $n$ great enough.
Thus, 
\begin{align} \label{eq:convergence-rate-E-epsilon}
    \left|\frac{1}{n}\sum_{h=-n}^n
    \sum_{i=n_{\epsilon,|h|}+1}^{n-|h|}(v(i,i+|h|)-1)C^X(|h|) \right|
    &\leq \frac{\epsilon}{n}\sum_{h=-n}^n(n-|h|-n_{\epsilon,|h|-1}-1) |C^X(|h|)|
    \nonumber
    \\
    &\leq \frac{\epsilon}{n}\sum_{h=-n}^nn |C^X(|h|)|
    \nonumber
    \\
    &=\epsilon\sum_{h=-n}^n |C^X(|h|)|.
\end{align}
We further observe
\begin{align*}
    \epsilon \sum_{h=-n}^n |C^X(|h|)|
    & \overset{n\to \infty}{\to}
    \epsilon \sum_{h\in \Z} |C^X(|h|)| 
    < \infty
\end{align*}
 by assumption (iii).
Since $\epsilon$ might be chosen arbitrarily small for great enough $n$, we obtain 
\begin{align*}
    \left|\frac{1}{n}\sum_{h=-n}^n
\sum_{i=n_{\epsilon,|h|}+1}^{n-|h|}(v(i,i+|h|)-1)C^X(|h|) \right|
&\overset{n\to \infty}{\to}
    0.
\end{align*}
At last, using assumption (ii) and (iv) we calculate
\begin{align}\label{eq:convergence-rate-E-n-epsilon}
    \left|\frac{1}{n}\sum_{h=-n}^{n}
\sum_{i=1}^{n_{\epsilon,|h|}}(v(i,i+|h|)-1)C^X(|h|) \right|
&\leq 
\frac{1}{n}\sum_{h=-n}^{n}
\sum_{i=1}^{n_{\epsilon,|h|}}|(v(i,i+|h|)-1)||C^X(|h|)|
\nonumber
\\
&\leq 
\frac{(C+1)}{n}\sum_{h=-n}^{n}
\sum_{i=1}^{n_{\epsilon,|h|}}|C^X(|h|)|
\nonumber
\\
&=
\frac{(C+1)}{n}\sum_{h=-n}^{n}
n_{\epsilon,|h|}|C^X(|h|)|
\\
&\overset{n\to \infty}{\to} 0 \tag*{\qedhere}
\end{align}
\end{proof}

\begin{lemma}\label{lem:variance}
    Assume the following:
    \begin{enumerate}[label=(\roman*)]
        \item $\E(X_i)=0$ for all $i$.
        \item $|v(i,j)|\leq C$ for some $C\in \R$ and all $i,j$.
        \item for all $i\in \N$ the sum $\sum_{h\in \Z}v(i,i+|h|)=C_i\in \R$ is finite and
        \item $\lim_{n\to \infty}\frac{1}{n^2}\sum_{i=1}^nC_i=0$
        \item $\E(X_i^{8})<\infty$ and $\sum_{i=1}^\infty\alpha(i)^{\frac{1}{2}}<\infty$
    \end{enumerate}
    where $\alpha$ is the dependence coefficient of the $\alpha$-mixing sequence $(X_i)_{i\in \N}$.
    
    Then, 
    $$\var^X\left[\var^V\left(\frac{1}{\sqrt{n}}\sum_{i=1}^n V_iX_i\right)\right]\le \mathcal{O}\left(\frac{1}{n^2}\sum_{i=1}^nC_i\right) \overset{n\to \infty}{\to}0.$$
\end{lemma}

\begin{proof}
    By \Cref{lem:mixing-cov-bound}, (v) implies
    \begin{align}\label{eq:sum-cov-bounded}
        \sum_{h=-\infty}^\infty C^X(|h|)=K\in \R.
    \end{align}
    Using (i), we calculate 
    \begin{align}\label{eq:Kunsch}
        \var^X\left[\var^V\left(\frac{1}{\sqrt{n}}\sum_{i=1}^nV_iX_i\right)\right]
        & = \frac{1}{n^2}\sum_{i_1,\dots,i_4=1}^nv(i_1,i_2)v(i_3,i_4)\cov(X_{i_1}X_{i_2},X_{i_3}X_{i_4}).
    \end{align}
    We consider the following term
    \begin{align*}
        \frac{1}{n^2}\sum_{i_1,\dots,i_4=1}^nv(i_1,i_2)v(i_3,i_4)(C^X(|i_1-i_3|)C^X(|i_2-i_4|)+C^X(|i_1-i_4|)C^X(|i_2-i_3|)).
    \end{align*}
    Substituting $i_2=i_1+s_1,i_3=i_1+s_2,i_4=i_1+s_3$
    and using \Cref{lem:Kuensch}, \eqref{eq:Kunsch} is asymptotically equivalent to 
    \begin{align}\label{eq:substituting}
        \frac{1}{n^2}\sum_{i_1=1}^n\sum_{s_1,s_2,s_3\in \Z}&(v(i_1,i_1+|s_1|)v(i_1+|s_2|,i_1+|s_3|)\nonumber\\
        &\cdot (C^X(|s_2|)C^X(|s_1-s_3|)+C^X(|s_3|)C^X(|s_1-s_2|))).
    \end{align}
    We observe further $\sum_{h\in \Z}C^X(|h-k|)=
        \sum_{h\in \Z}C^X(|h|)$
    for all $k\in \Z$ and, hence,
    \begin{align*}
        \sum_{s_1,s_3\in \Z}v(i_1,i_1+|s_1|) C^X(|s_1-s_3|)
        &=\sum_{s_1\in \Z}v(i_1,i_1+|s_1|) \sum_{s_3\in \Z}C^X(|s_1-s_3|)
        \\
        &=\sum_{s_1\in \Z}v(i_1,i_1+|s_1|) \sum_{s_3\in \Z}C^X(|s_3|)
        \\
        &=C_{i_1}K,
    \end{align*}
    by (iii) and (vi) and similarly
    \begin{align*}
        \sum_{s_1,s_2\in \Z}v(i_1,i_1+|s_1|) C^X(|s_1-s_2|)
        &=C_{i_1}K.
    \end{align*}
    Combined with (ii), we obtain
    \begin{align}\label{eq:convergence-rate-variance}
        \eqref{eq:substituting}
        \le \frac{1}{n^2}\sum_{i_1=1}^n2C_{i_1}CK^2
        \stackrel{n\to \infty}{\to}0
        \qquad \text{by (iv).}
    \end{align}
    This establishes the claim.
\end{proof}

\begin{lemma}\label{cor:var-convergence}
    Suppose the following conditions hold:
    \begin{enumerate}[label=(\roman*)]
        \item $\E(X_i)=0$ and $\E(X_i^8) < \infty$ for all $i$ and $\sum_{h=-\infty}^\infty |C^X(|h|)|<\infty$. 
        \item  $\sum_{i=1}^\infty\alpha(i)^{\frac{1}{2}}<\infty$.
        \item $|v(i,j)|\leq C$ for some $C\in \R$ and all $i,j$.
        \item for all $\epsilon>0$ and $h$ such that $C^X(|h|)\neq 0$ there exists $n_{\epsilon,|h|}$ such that $$|1-v(i,i+|h|)| \le \epsilon$$ for all $i \ge n_{\epsilon,|h|}$ and $\lim_{n\to \infty}\frac{1}{n}\sum_{h=-n}^n n_{\epsilon,|h|} |C^X(|h|)|=0$
        \item for all $i\in \N$ the sum $\sum_{h\in \Z}v(i,i+|h|)=C_i\in \R$ is finite and $\lim_{n\to \infty}\frac{1}{n^2}\sum_{i=1}^nC_i=0$.
    \end{enumerate}
    Then $$\var^V\left(\frac{1}{\sqrt{n}}\sum_{i=1}^nV_i(X_i-\E(X_1))\right)\overset{P}{\to}\sigma_\infty^2.$$
\end{lemma}
\begin{proof}
    Follows immediately from  \cref{lem:expected-value}, \cref{lem:variance}, and Chebyshev's inequality.
\end{proof}

\begin{lemma}\label{lem:ass-cons-var-example}
    Let $V_i$ be as in \Cref{ex:construction}.
    Then, 
    \begin{enumerate}
        \item $V_i\sim \mathcal{N}(1,1)$ for all $i$
        \item for all $h\in \Z$ and $i\in \N$ it holds $$0\le \cov(V_i,V_{i+|h|})= \prod_{k=1}^{|h|}(1-(i+k)^{-\beta}) \le 1$$
        \item for all $\epsilon<1$, $h\in \Z$ and $n_{\epsilon,|h|}=(1-\sqrt[|h|]{1-\epsilon})^{-\frac{1}{\beta}}$ it holds $$|1-\cov(V_i,V_{i+|h|})|\le \epsilon$$ for all $i\ge n_{\epsilon,|h|}$
        \item $C_i=\sum_{h\in \Z}\cov(V_i,V_{i+|h|})\le 4i^\beta+B$ for some constant $B\in \R$ independent of $i$ and, in particular, $$\lim_{n\to \infty}\frac{1}{n^2}\sum_{i=1}^nC_i=0.$$
    \end{enumerate}
    \vspace*{-18pt}
\end{lemma}
\begin{proof}
    Observe all $V_i$ to be normally distributed as sums of normally distributed random variables.
    Inductively, we obtain $\var(V_i)=1=\E(V_i)$.
    This proves 1.

    We calculate 
    \begin{align*}
        0&\le 
        \cov(V_i,V_{i+|h|})
        \\
        &=\rho_{i+|h|}\cov(V_i,V_{i+|h|-1})
        \\
        &=\prod_{k=1}^{|h|}\rho_{i+k}\var(V_i)
        &\text{inductively}
        \\
        &=\prod_{k=1}^{|h|}\rho_{i+k}
        \\
        &=\prod_{k=1}^{|h|}(1-(i+k)^{-\beta})
        \\
        &\leq (1-(i+|h|)^{-\beta})^{|h|}
    \end{align*} for $h\in \Z$ and, in particular, $$|\cov(V_i,V_{i+h})|\leq 1$$ for all $i,h$ which proves 2.
    
    Furthermore, 
    \begin{align*}
        |1-\cov(V_i,V_{i+|h|})|
        =1-\prod_{k=1}^{|h|}(1-(i+k)^{-\beta})
        \le 1-(1-i^{-\beta})^{|h|}.
    \end{align*}
    For any $\epsilon \in (0, 1)$, we calculate
    \begin{align*}
        (1-(1-i^{-\beta})^{|h|})\le\epsilon
        \qquad \Leftrightarrow \qquad 
        1-i^{-\beta} \ge\sqrt[|h|]{1-\epsilon}
        \qquad 
        \Leftrightarrow \qquad 
        i \ge(1-\sqrt[|h|]{1-\epsilon})^{-\frac{1}{\beta}}.
    \end{align*}
    Hence, we obtain 3.

    Furthermore,
    \begin{align*}
        \sum_{h\in \Z}\cov(V_i,V_{i+|h|})
        &\leq 2\sum_{h=0}^\infty(1-(i+h)^{-\beta})^h 
        \\
        &=2\left[\sum_{h=0}^i(1-(i+h)^{-\beta})^h+\sum_{h=i+1}^\infty(1-(i+h)^{-\beta})^h \right]
        \\
        &\leq 2\left[\sum_{h=0}^\infty(1-(2i)^{-\beta})^h+\sum_{h=0}^\infty(1-(2h)^{-\beta})^h \right]
        \\
        &= 2\left[(2i)^\beta+\sum_{h=0}^\infty(1-(2h)^{-\beta})^h \right]
        &\text{by the gemoetric series}
        \\
        &=C_i< \infty,
    \end{align*}
    where the last step follows from $$0\leq \sum_{h=0}^\infty(1-(2h)^{-\beta})^h<\infty$$ 
    for all $\beta\in (0,1)$.
    Thus, we calculate 
    \begin{align}\label{eq:example-online-sequence-Ci}
        0\le\lim_{n\to \infty}\frac{1}{n^2}\sum_{i=1}^nC_i=
    \lim_{n\to \infty}\frac{4}{n^2}\sum_{i=1}^ni^{\beta}\leq \lim_{n\to \infty}\frac{4}{n}n^{\beta}=0
    \end{align}
    and, hence, we obtain 4.
    \end{proof}
    
    \begin{proof}[Proof of \Cref*{thm:variance-details}]
    To prove the parts (a)--(c) of the theorem 
    it suffices to check the conditions of \cref{cor:var-convergence}, which also imply the conditions of \crefrange{lem:expected-value}{cor:var-convergence}.
    Conditions (i) and (ii) are given by assumption of the theorem.
    Conditions (iii)--(v) were proven in \Cref{lem:ass-cons-var-example}.

    For (iv) we calculated $n_{\epsilon,|h|}=(1-\sqrt[|h|]{1-\epsilon})^{-\frac{1}{\beta}}$ in 
    \Cref{lem:ass-cons-var-example}.
    We approximate $$f(\epsilon)=1-\sqrt[|h|]{1-\epsilon}\approx \frac{\epsilon}{|h|}$$ through a first order Taylor series around $\epsilon=0$.
    Thus, we calculate
    $$n_{\epsilon,|h|}=(1-\sqrt[|h|]{1-\epsilon})^{-\frac{1}{\beta}}\approx \biggl(\frac{|h|}{\epsilon}\biggr)^{\frac{1}{\beta}}=\mathcal{O}(|h|^{\frac{1}{\beta}}).$$
    In order to determine an optimal $\beta$ we minimize the MSE
    \begin{align*}
        &\quad \E^X \left[ \left(\var^V\left(\frac{1}{\sqrt{n}}\sum_{i=1}^n V_iX_i\right) -\sigma_\infty^2\right)^2\right] \\
        &=\E^X\left[\var^V\left(\frac{1}{\sqrt{n}}\sum_{i=1}^n V_iX_i \right)-\sigma_\infty^2\right]^2
        +\var^X\left[\var^V\left(\frac{1}{\sqrt{n}}\sum_{i=1}^n V_iX_i\right)\right]
    \end{align*}
    for $n\to \infty$.
    From \eqref{eq:convergence-rate-variance} and \eqref{eq:example-online-sequence-Ci}, we get 
    $$\var^X\left[\var^V\left(\frac{1}{\sqrt{n}}\sum_{i=1}^n V_iX_i\right)\right] = \Ocal\biggl(\frac{1}{n^2}\sum_{i=1}^nC_i\biggr)=\mathcal{O}(n^{\beta-1}).$$
    
    From \eqref{eq:convergence-rate-E-epsilon} and \eqref{eq:convergence-rate-E-n-epsilon}, we get 
    $$\E^X\biggr[\var^V\left(\frac{1}{\sqrt{n}}\sum_{i=1}^n V_iX_i\right)-\sigma_\infty^2\biggr] = \Ocal\left(\frac{1}{n}\sum_{h=-n}^nn_{\epsilon,|h|}|C^X(|h|)| + \epsilon\sum_{h=-n}^n|C^X(|h|)| \right).$$
    We determine a sequence of $\epsilon_n\to 0$ in order to balance both.
    
    Assuming $$A=\sum_{h\in \Z}|C^X(|h|)|,B=\sum_{h\in \Z}|h|^{\frac{1}{\beta}}C^X(|h|)<\infty.$$
    Then, $\epsilon_n$ minimizing the sum of $$\frac{1}{n}\sum_{h=-n}^nn_{\epsilon,|h|}|C^X(|h|)| \text{ and }\epsilon\sum_{h=-n}^n|C^X(|h|)|$$ is (asymptotically) given by 
    $$\epsilon_n=\min_{\epsilon>0}\biggl(\frac{1}{n}B\epsilon^{-\frac{1}{\beta}} +A\epsilon\biggr).$$
    Calculating the derivative with respect to $\epsilon$ yields 
    $$-\frac{1}{\beta n}B\epsilon^{-\frac{\beta+1}{\beta}} +A$$
    whose only root is given by $$\epsilon_n=\left(\frac{A}{B}\beta n\right)^{-\frac{\beta}{1+\beta}}.$$
    
    A straightforward calculation shows 
    $$\frac{1}{n}\epsilon_n^{-\frac{1}{\beta}} 
    +\epsilon_n=n^{-\frac{1+\beta}{1+\beta}}\left(\frac{A}{B}\beta n\right)^{\frac{1}{1+\beta}}+\epsilon_n=2\epsilon_n.$$
    Combined, we obtain the convergence rate 
    $$\E^X \left[ \var^V\left(\frac{1}{\sqrt{n}}\sum_{i=1}^n V_iX_i\right) -\sigma_\infty^2\right] =\mathcal{O}\left(\frac{1}{n}\epsilon_n^{-\frac{1}{\beta}} +\epsilon_n\right)=\mathcal{O}(2\epsilon_n)=\mathcal{O}(n^{-\frac{\beta}{1+\beta}}).$$

    Finally, we pick $\beta$ to minimize the MSE
    \begin{align*}
        E^X \left[ \left(\var^V\left(\frac{1}{\sqrt{n}}\sum_{i=1}^n V_iX_i\right) -\sigma_\infty^2\right)^2\right]
        =&\mathcal{O}\left(n^{-\frac{2\beta}{1+\beta}}\right)+\mathcal{O}(n^{\beta-1}).
    \end{align*}
    Minimizing
    $$n^{-\frac{2\beta}{1+\beta}}+n^{\beta-1}$$
    for $n\to \infty$ amounts in minimizing the sum of exponents
    $$-\frac{2\beta}{1+\beta}+(\beta-1)=\frac{\beta^2-2\beta-1}{1+\beta}.$$
    Its derivative is given by $$\frac{(\beta+1)^2-2}{(\beta+1)^2}.$$
    Thus the optimal $\beta\in (0,\frac{1}{2})$ is given by $\beta_{opt}=\sqrt{2}-1.$
\end{proof}

\section{Central limit theorem}

\begin{proof}[Proof of \cref{thm:sequence-consistent}]

Theorem 1.7 in \cite{bosq2012nonparametric}
with $\gamma=4$ implies that 
$$\frac{1}{\sqrt{n}}\sum_{i=1}^n(X_i-\E(X_i)) \stackrel{d}\to \Ncal(0, \sigma_\infty^2),$$
as $n \to \infty$, or, written differently,
\begin{align}
    \sup_{x\in \R}&\bigl|\Pr\left\{\sqrt{n}(T_n-\E(T_n)) \leq x \right\} - \Phi(x/\sigma_\infty)\bigr| \to 0, \label{eq:x-clt}
\end{align}
where $T_n = n^{-1} \sum_{i = 1}^n X_i$
and $\Phi$ the standard normal cumulative distribution function.
It remains to show that the centered bootstrap average has the same limiting distribution.
Recall from \cref{lem:vn-negligible} that 
\begin{align*}
    &\quad \sqrt{n}(T_n^* - T_n) = \frac{1}{\sqrt{n}}\sum_{i = 1}^n \frac{V_i}{\bar V_n} X_i - \frac{1}{\sqrt{n}} \sum_{i = 1}^n X_i  
    = \frac{1}{\sqrt{n}}\sum_{i = 1}^n (V_i - 1) (X_i - \E[X_i]) + o_P(1).
\end{align*}
The distribution of the sum on the right, conditional on $X_1, \dots, X_n$, is normal with mean 0 and variance
\begin{align*}
    \sigma_n^2 := \var^V\left(\frac{1}{\sqrt{n}}\sum_{i = 1}^n (V_i - 1) (X_i - \E[X_i])\right) = \var^V\left(\frac{1}{\sqrt{n}}\sum_{i = 1}^n V_i (X_i - \E[X_i])\right),
\end{align*}
which converges in probability to $\sigma_\infty^2$ by \cref{cor:var-convergence}.
We therefore get
\begin{align*}
     \sup_{x \in \R}|\Pr^*\{\sqrt{n}(T_n^* - T_n) \le x\} - \Phi(x/\sigma_\infty) |  
    &\le \sup_{x \in \R}|\Pr^*\{\sqrt{n}(T_n^* - T_n) \le x\} - \Phi(x/\sigma_n) | + \sup_{x \in \R}| \Phi(x/\sigma_n) - \Phi(x/\sigma_\infty) | \\&\stackrel{P}\to 0,
\end{align*}
where the rightmost supremum converges by the continuous mapping theorem.
Together with \eqref{eq:x-clt}, we obtain
\begin{align*}
    &\quad \sup_{x \in \R}|\Pr^*\{\sqrt{n}(T_n^* - T_n) \le x\} - \Pr\left\{\sqrt{n}(T_n-\E(T_n)) \leq x \right\} | \\
    &\le \sup_{x \in \R}|\Pr^*\{\sqrt{n}(T_n^* - T_n) \le x\} - \Phi(x/\sigma_\infty)| + \sup_{x \in \R}|\Pr\left\{\sqrt{n}(T_n-\E(T_n)) \leq x \right\} - \Phi(x/\sigma_\infty) | \stackrel{P}\to 0. \tag*{\qedhere}
\end{align*}

\end{proof}

\section{Details on Moving average block bootstrap}
\begin{example}[Multiplier block bootstrap with MA-weights]\label{ex:ma-bootstrap-buehlman}
    Consider the moving average (MA) process 
    $$V_{j, n}=\sum_{j\in \Z} b_{j,n}\zeta_{i-j,n}$$ 
    with 
    \[
    b_j = 
        \begin{cases}
            m_n^{-1}(1-|j|/m_n) & |j|\leq m_n\\
            0 & \text{else,}
        \end{cases}
    \]
    where $\zeta_{i,n}\stackrel{iid}\sim Gamma(q_n,q_n)$ with $q_n=\frac{2}{3m_n}+\frac{1}{3m_n^2}$
    and $m_n\sim Cn^{1/3}$ asymptotically.    
    Define the resampling scheme $X_{i,n}^*=X_iV_{i,n}/\bar{V}_{n}$.
\end{example}

\section{Evaluation metrics}\label{ap:evaluation-metrics}
Recall that \Cref{thm:variance-details} yields a consistent variance estimator,
i.e.
\begin{align*}
    \var^V\left(\frac{1}{\sqrt{n}}\sum_{i=1}^n\frac{V_i}{\bar{V}_n}X_i\right)=\var^V\left(\frac{1}{\sqrt{n}}\sum_{i=1}^n\left(\frac{V_i}{\bar{V}_n}-1\right)X_i\right)&\to \sigma_\infty 
\end{align*}
in probability.
Furthermore,
\Cref{thm:sequence-consistent} yields asymptotically consistent confidence 
intervals $C_n$ at level $\alpha$ for $\theta=\E(X_1)=\E(X_i)$ by Lemma 23.3 of \cite{van2000asymptotic}, i.e.
$$\lim\inf_{n\to \infty}P(\theta\in C_n(X_1,...,X_n))\geq 1-\alpha.$$

Accordingly, given some realizations $x_1,...,x_n$, we obtain
\begin{align*}
    ev_1(x_1,...,x_n)
    &=\var^V\left(\frac{1}{\sqrt{n}}\sum_{i=1}^n\frac{V_i}{\bar{V}_n}x_i\right) 
    \\
    ev_2(x_1,...,x_n)
    &=\begin{cases} 
            1 & \theta\in C_n(x_1,...,x_n)  \\
            0 & \theta\notin C_n(x_1,...,x_n)
        \end{cases}
\end{align*}
which provide quantities of the accuracy of the bootstrap procedure.
Observe $$\E(ev_2(X_1,...,X_n))=P(\theta\in C_n(X_1,...,X_n))$$
and $$ev_1(X_1,...,X_n)=\var^V\left(\frac{1}{\sqrt{n}}\sum_{i=1}^n\frac{V_i}{\bar{V}_n}X_i\right)$$
by the very construction.

Following standard practice, 
we estimate the
mean $E_1$ and variance $E_2$ of $ev_1(X_1,...,X_n)$ as well as 
the expected value $E_3$ of $ev_2(X_1,...,X_n)$
by sampling $M=250$ times from the 
underlying bootstrap distribution, i.e. realize $v_{1,j},....v_{n,j}\in \R$ according 
$v_{i,j}\sim V_i$ for 
$j=1,...,M$ and calculate
\begin{align*}
    \hat{ev}_1(x_1,...,x_n)
    &=\frac{1}{M-1}\sum_{j=1}^M\left(\frac{1}{\sqrt{n}}\sum_{i=1}^n\frac{v_{i,j}}{\bar{v_n}}x_i\right)^2 
    \\
    \hat{ev}_2(x_1,...,x_n)
    &=\begin{cases} 
        1 & \theta\in \hat{C}_n(x_1,...,x_n)  \\
        0 & \theta\notin \hat{C}_n(x_1,...,x_n)
    \end{cases}
\end{align*}
where $\hat{C}_n(x_1,...,x_n)$ denotes corresponding estimate of $C_n(x_1,...,x_n)$.

\section{Auxilary results}

\begin{lemma}\label{lem:mixing-cov-bound}
    Assume $$\E(X_i^4)<\infty$$ and $$\sum_{i=1}^\infty\alpha(i)^{\frac{1}{2}}<\infty.$$
    Then, $$\sum_{h=-\infty}^\infty |C^X(|h|)|<\infty.$$
\end{lemma}

\begin{proof}
    We apply (corollary 1.1 Bosq) to $p=\frac{1}{2}$ and $r=q=\frac{1}{4}$.
    A straightforward calculation exhibits 
    $$\frac{1}{q}+\frac{1}{r}=1-\frac{1}{p}.$$
    We obtain 
    $$|C^X(|h|)|\le 2p\alpha(|h|)^{\frac{1}{2}}\E(X_1^4)^{\frac{1}{2}}$$
    by stationarity of $(X_i)_{i\in \N}$.
    Hence, $$\sum_{h=-\infty}^\infty |C^X(|h|)|<\infty$$ follows from $$\sum_{i=1}^\infty\alpha(i)^{\frac{1}{2}}<\infty.$$
\end{proof}

\begin{lemma}\label{lem:Kuensch}
    Let $(X_i)_{i\in \N}$ be a zero mean, real valued, strictly stationary strong mixing stochastic process
    with eigth moments existing.
    
    Then, for all $a,b,c,d\in \N$
    \begin{align*}
        &|\E(X_aX_bX_cX_d)-\E(X_aX_b)\E(X_cX_d)-\E(X_aX_c)\E(X_bX_d)-\E(X_aX_d)\E(X_bX_c)|
        \\ &\le C \alpha(\max\{i_2-i_1,i_3-i_2,i_4-i_3\})^{\frac{1}{2}}
    \end{align*}
    with $\{a,b,c,d\}=\{i_1,i_2,i_3,i_4\}$ such that $i_1\leq i_2\leq i_3\leq i_4$ 
    and $C\in \R$ independent of $a,b,c,d\in \N$.
    
    In particular, fixing $a$, we obtain
    $$\E(X_aX_bX_cX_d)-\E(X_aX_b)\E(X_cX_d)=\E(X_aX_c)\E(X_bX_d)-\E(X_aX_d)\E(X_bX_c)+\epsilon_{b,c,d}$$
    with $\epsilon_{b,c,d}\to 0$ if some $b,c$ or $d$ tends to infinity
\end{lemma}

\begin{proof}
   In order to remain clarity, we only outline the proof rather than going into 
   detaill. 

   Since 
   $$|\E(X_aX_bX_cX_d)-\E(X_aX_b)\E(X_cX_d)-\E(X_aX_c)\E(X_bX_d)-\E(X_aX_d)\E(X_bX_c)|$$
   is invariant under permutation of the $a,b,c,d$, we may without loss of generality assume 
   $a\leq b\leq c\leq d$, i.e. $a=i_1,...,d=i_4$.
   Then, we obtain
   \begin{align*}
    |\E(X_aX_bX_cX_d)-\E(X_aX_b)\E(X_cX_d)|=|\cov(X_aX_b,X_cX_d)|&\leq K \alpha(c-b)^{1/2}\E(X_1^8)^{1/2}\\
    |\E(X_aX_c)|= |\cov(X_a,X_c)|&\leq K \alpha(c-a)^{1/2}\E(X_1^4)^{1/2} \\
    |\E(X_bX_c)|= |\cov(X_b,X_c)|&\leq K \alpha(c-b)^{1/2}\E(X_1^4)^{1/2}
   \end{align*}
   by applying Corollary 1.1 in \citet{bosq2012nonparametric}
   with $q=r=4,p=2$ to the respective covariance and by using Cauchy-Schwarz inequality.
   Combined, triangle inequality yields
   \begin{align*}
        &|\E(X_aX_bX_cX_d)-\E(X_aX_b)\E(X_cX_d)-\E(X_aX_c)\E(X_bX_d)-\E(X_aX_d)\E(X_bX_c)|
        \\ &\le C \alpha(c-b)^{\frac{1}{2}}.
    \end{align*}
    for some $C$ independent of $a,...,d$.
    By the very same argument we derive 
    \begin{align*}
        |\E(X_aX_b)|= |\cov(X_a,X_c)|&\leq K \alpha(b-a)^{1/2}\E(X_1^4)^{1/2} \\
        |\E(X_aX_c)|= |\cov(X_b,X_c)|&\leq K \alpha(b-a)^{1/2}\E(X_1^4)^{1/2} \\
        |\E(X_aX_d)|= |\cov(X_b,X_c)|&\leq K \alpha(b-a)^{1/2}\E(X_1^4)^{1/2}
    \end{align*}
    Next, we apply Corollary 1.1 in \citet{bosq2012nonparametric} with $q=8/3,r=8$ and $p=2$
    in order to obtain
    \begin{align*}
        |\E(X_aX_bX_cX_d)|=|\cov(X_a,X_bX_cX_d)|&\leq K \alpha(b-a)^{1/2}\E(X_1^8)^{1/8}\E((X_bX_cX_d)^{8/3})^{3/8}\\
        &\leq K \alpha(b-a)^{1/2}\E(X_1^8)^{1/2}
    \end{align*}
    where we applied Hölder's inequality in the last step.
    Thus, triangle inequality yields 
    \begin{align*}
        &|\E(X_aX_bX_cX_d)-\E(X_aX_b)\E(X_cX_d)-\E(X_aX_c)\E(X_bX_d)-\E(X_aX_d)\E(X_bX_c)|
        \\ &\le C \alpha(b-a)^{\frac{1}{2}}
    \end{align*}
    and the same argument yields 
    \begin{align*}
        &|\E(X_aX_bX_cX_d)-\E(X_aX_b)\E(X_cX_d)-\E(X_aX_c)\E(X_bX_d)-\E(X_aX_d)\E(X_bX_c)|
        \\ &\le C \alpha(d-c)^{\frac{1}{2}}
    \end{align*}
    for some constant independent of $a,...,d$.
    This proves the first claim.

    The last claim follows by the strong mixing property of $(X_i)$.
    This completes the proof.
\end{proof}